 \documentclass[letterpaper, 10 pt, conference]{ieeeconf}  

\IEEEoverridecommandlockouts                              

\usepackage{graphics} 
\usepackage{epsfig} 
\usepackage{mathptmx} 
\usepackage{times} 
\usepackage{amsmath} 
\usepackage{amssymb}  
\usepackage{bm}
\usepackage{algorithm}
\usepackage{multirow}
\usepackage{verbatim}
\usepackage{graphicx}
\usepackage{subfigure}
\usepackage{color}
\usepackage{url}

\title{\LARGE \bf
Assistive arm and hand manipulation: How does current research intersect with actual healthcare needs?
}

\author{Laura Petrich$^{1}$, Jun Jin$^{1}$, Masood Dehghan$^{1}$ and Martin Jagersand$^{1}$ 
\thanks{$^{1}$L. Petrich, J. Jin, M. Dehghan and M. Jagersand are with Department of Computing Science, 
        University of Alberta, Canada,
        {\tt\small jag@cs.ualberta.ca}}%
}

\begin{document}

\maketitle
\thispagestyle{empty}
\pagestyle{empty}


\begin{abstract}
Human assistive robotics have the potential to help the elderly and individuals living with disabilities with their Activities of Daily Living (ADL). Robotics researchers present bottom up solutions using various control methods for different types of movements. Health research on the other hand focuses on clinical assessment and rehabilitation leaving arguably important differences between the two domains. In particular, little is known quantitatively on what ADLs humans perform in their everyday environment - at home, work etc. This information can help guide development and prioritization of robotic technology for in-home assistive robotic deployment. This study targets several lifelogging databases, where we compute (i) ADL task frequency from long-term low sampling frequency video and Internet of Things (IoT) sensor data, and (ii) short term arm and hand movement data from 30 fps video data of domestic tasks. Robotics and health care communities have different terms and taxonomies for representing tasks and motions. We derive and discuss a robotics-relevant taxonomy from this quantitative ADL task and ICF motion data in attempt to ameliorate these taxonomic differences. Our statistics quantify that humans reach, open drawers, doors, and retrieve and use objects hundreds of times a day. Commercial wheelchair mounted robot arms can help 150,000 upper body disabled in the USA alone, but only a few hundred robots are deployed. Better user interfaces, and more capable robots can increase the potential user base and number of ADL tasks solved significantly.
\end{abstract}

\section{Introduction}
        \label{sec:introduction}
Activities of Daily Living (ADL) can be a challenge for individuals living with upper limb disabilities and assistive robotics has the potential to help increase functional independence \cite{Smarr201026Environment}. Assistive robot arms, such as, the wheelchair-mountable Kinova Jaco \cite{Archambault201161Arm} and  Manus/iArm \cite{DriessenMANUSRobot}, have been commercially available for over a decade. Such devices can increase independence, decrease the caregiver load and reduce health care costs \cite{Stuyt200564A}. Robot arms have the potential to be as important to individuals living with upper body disabilities as power wheelchairs have become to those with lower body disabilities. However, outside of research purposes, only a few hundred assistive arms, primarily in Europe and North America, are practically deployed and in use. For assistive robotics research, knowing which ADLs are most important to support as well as the necessary performance parameters for these tasks will be crucial to increase usability and deployment. In order to build a task taxonomy that focuses on patient independence, it is also imperative to understand how the health care community defines independence. We briefly review the World Health Organization Disability Assessment Schedule (WHODAS2.0) \cite{Ustun2010}, and classifications of tasks and motions in the International Classification of Functioning (ICF) \cite{WorldHealthOrganizationWHO200339Health}. These were primarily developed to determine the level of disability and design corresponding rehabilitation plans, not to guide assistive robotics research. To the best of the authors knowledge, in health care literature there does not appear to be quantitative studies and statistics on able or disabled human use of ADL tasks and ICF motions.
This paper hopes to mitigate this quantitative gap.
\begin{figure}
    \begin{center}
        \includegraphics[width=0.51\textwidth]{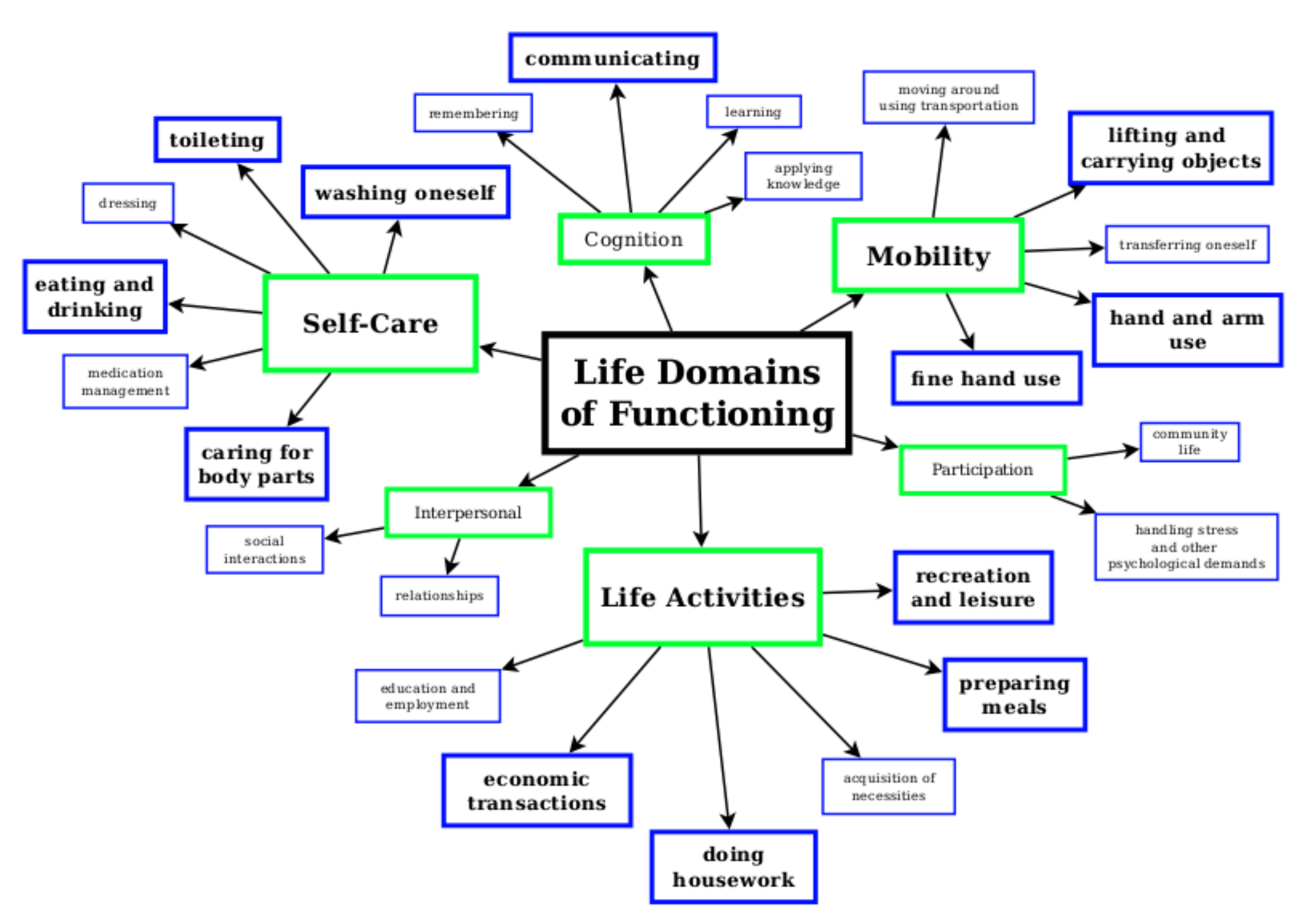} 
        \caption{The major life domains of functioning and disability as set out in the World Health Organization Disability Schedule 2.0 (WHODAS2.0); a standardized cross-cultural measurement of health status based on the International Classification of Functioning, Disability and Health. WHODAS2.0 can be used to measure the impact of health conditions, monitor intervention effectiveness and estimate the burden of physical and mental disorders across all major life domains.}
        \label{fig:LifeDomains} 
    \end{center}
\end{figure}
The use of robotics to help increase functional independence in individuals living with upper limb disabilities has been studied since the 1960’s. We distinguish here between a physically independent robot arm, typically mounted on the users wheelchair, and a smart prosthesis, attached to an amputated limb, with the former being our group of interest. 
The United States Veterans Affairs estimate that approximately 150,000 Americans could benefit from currently commercialy available wheelchair mounted robot arms\cite{Chung201316Review}. With improved functionality, reliability and ease of use, deployment to larger populations could be possible. 

In the field of Computer Science, recent interest in video object and activity recognition \cite{Wang201319Activities, Wang2016} along with life-logging capture has resulted in numerous public data-sets \cite{GurrinCathalJohoHideoHopfgartnerFrankZhouLitingAlbatal201623Research}. In this work we analyzed over 30 such data-sets in order to extract everyday tasks of high importance and relevant motion data \cite{Bolanos201725Overview}.

Health care and robotic domains use different taxonomies to classify and everyday activity tasks and motions \cite{Katz197934Activities,Lawton196936Living,Langdon201604Taxonomy,Bloomfield200332Tasks,Dollar2014ClassifyingLiving,Bullock201105Behavior}. By merging these taxonomies and quantifying health care needs with robotic capabilities we seek to bridge the two, often separate, communities. This would provide the robotics community with guidance as to which tasks would make a large impact to patient populations if implemented on assistive robotic systems.

\section{Activities of Daily Living, Self-Care, and Functional Independence}
        \label{sec:adl}
\begin{figure}
    \begin{center}
        \includegraphics[width=0.35\textwidth]{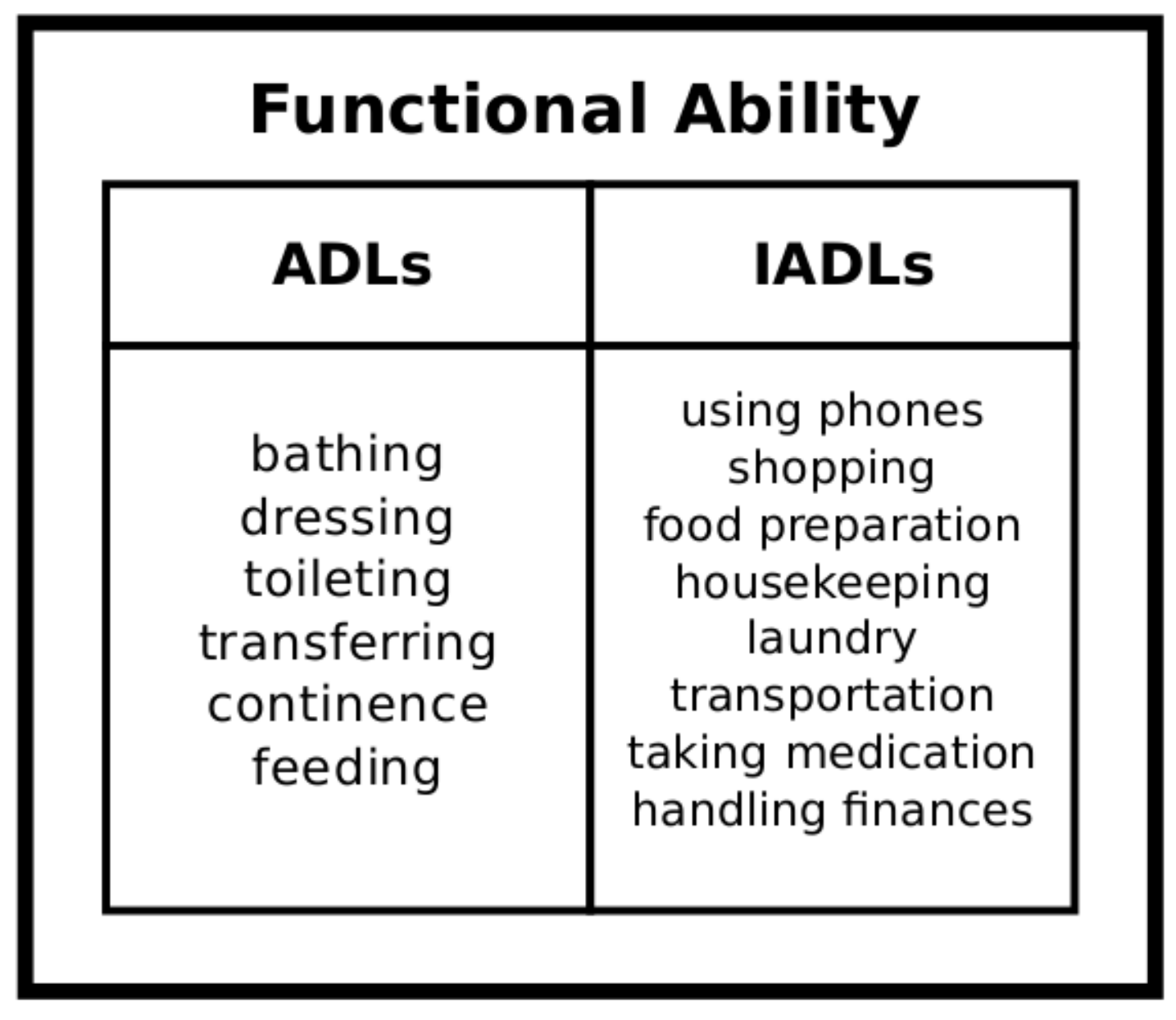} 
        \caption{Measures of Functional Ability: Activities of Daily Living (ADLs) and Instrumental Activities of Daily Living (IADLs). ADLs are basic self-care tasks whereas IADLs are more complex skills; together they represent what a person needs to be able to manage on their own in order to live independently or functional ability.}
        \label{fig:FunctionalAbility} 
    \end{center}
\end{figure}
The International Classification of Functioning, Disability and Health (ICF) provides a framework for determining the overall health of individuals and populations \cite{WorldHealthOrganizationWHO200339Health}. Disability information is an important indicator of a population’s health status, as it shows the impact that functional limitations have on independence. This concept is known as functional disability, or the limitations one may experience in performing independent living tasks \cite{Spector199837Disability}. A quantification of functional disability includes both measures of Activities of Daily Living (ADLs) \cite{Katz197934Activities} and Instrumental Activities of Daily Living (IADLs) \cite{Lawton196936Living} as shown in Figure \ref{fig:FunctionalAbility}; in this work we will refer to these collectively as ADLs. The World Health Organization (WHO) further developed WHODAS2.0 from ICF as a standardized, cross-cultural measure of functioning and disability across all life domains \cite{Federici201709Review}, see Fig. \ref{fig:LifeDomains}. 

A common approach that drives research is to ask patients and caregivers (nurses or family/friends) for their preferences when it comes to robotic assistance \cite{Beer201213Place,Stanger199427Priorities}. Notably, preferences vary and users opinions shift over time. In particular, caregivers tend to favor essential tasks such as taking medication, while patients favor picking up dropped objects and leisure related tasks in pre-automation surveys. When 67 users were surveyed both before and after they received and used an assistive robotic arm it was found that post-automation user preferences had shifted from leisure to work related tasks  \cite{Chung201316Review}. Comparing preferences with quantitative ADL data (Section \ref{sec:lifelogging}) will be important for health care priority decisions.

\begin{figure}
    \begin{center}
        \includegraphics[width=0.5\textwidth]{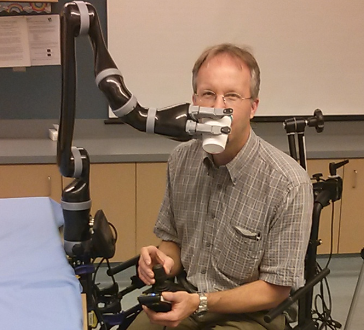} 
        \caption{Jaco assistive arm at the Glenrose Rehabilitation Hospital, Edmonton, Alberta.}
        \label{fig:Jaco} 
    \end{center}
\end{figure}
\section{A Review of Robotics for Assisted Living}
        \label{sec:robotic_review}
A lightweight robotic arm can be attached to a wheelchair to assist with ADLs \cite{Chung201316Review}. With such a device, users with limited upper limb functionality are able to independently carry out a larger subset of their daily tasks. While there are about 2 million robot arms deployed in industry, only two assistive robot manufacturers have over 100 assistive arms deployed with disabled humans, namely, Exact Dynamics (Manus and iARM) \cite{DriessenMANUSRobot} and Kinova (JACO and MICO) \cite{Campeau-lecours2018KinovaApplications}. These arms are lightweight with integrated controllers and cost around USD 20,000-35,000 with a gripper. For example, the Kinova JACO robotic arm weighs 5.7kg (12.5lbs) and comes with a wheelchair attachment kit. It is capable of grasping and moving objects up to 1.5kg, Fig.~\ref{fig:Jaco}. The Manus/iARM has similar specifications. 

In published assistive robotics research a variety of commercial robot arms are used and several new prototype arms have been designed, however, neither new robots nor new methods for motion control or Human Robot Interaction (HRI) have reached noticeable deployment \cite{Groothuis201311Arms}. The few hundred deployed JACO and Manus arms still use basic joystick position-based teleoperation, where a 2 or 3 Degree of Freedom (DOF) joystick is mapped to a subset of the Cartesian arm translation and wrist rotation controls \cite{Archambault201161Arm, Goodrich200706bSurvey}. To complete 6-DOF tasks the user needs to switch between various Cartesian planes, known as mode switching, which can be tedious and cognitively taxing.

Novel user interfaces have been implemented in research settings and rely on a variety of input signals for shared autonomy, such as gestures, eye gaze, electromyography (EMG), electroencephalography (EEG), and electrocorticographic (ECoG). Gesture-based systems allow the user to specify an object to manipulate by either pointing \cite{QuinteroRJ15} or clicking on it through a touch screen interface \cite{Jagersand1995,Gridseth2016} and then the robotic arm would autonomously move towards the target object \cite{Tsui201153Arm}. Eye gaze can be used in place of gestures to capture an individuals intended actions and drive robot control \cite{Admoni2016}. Neural interface systems (i.e. ECoG and EEG) work by mapping neuronal activity to multidimensional control signals that are used to drive robot arm movements \cite{Hochberg2013}. Hochberg \textit{et al.} highlight the potential of ECoG-based control methods, although it requires an invasive surgical procedure in order to implant the microelectrode array. EEG- and EMG-based methods provide an intuitive, non-invasive alternative for closed-loop robot control using brain and muscle activity \cite{Gomez2017, Liarokapis2012}. Recently, Struijk \textit{et al.} developed a wireless intraoral tongue interface device that enables individuals with tetraplegia to use their tongue to control a 7-DOF robotic arm and 3 finger gripper \cite{Struijk2017}.

\section{A Task Taxonomy for Arm Manipulation}
        \label{sec:taxonomy}
Robotic capabilities are built bottom-up by designing control methods for individual motions (i.e. motor primitives) which can be combined to solve specific tasks \cite{Kober2009}. The same motions can potentially be used to solve different ADLs that fall within the healthcare taxonomies. Dexterous in-hand manipulation requires different contact configurations and manipulation taxonomies have been developed to compensate for these various configurations  \cite{Bullock201305bManipulationb}. Robot arm manipulation is generally thought of as a 6-DOF Euclidean (end-effector) transform, thus requiring no taxonomy. Contrarily, ADL tasks naturally contain a variety of movements with different DOFs, as well as, contact and precision motions. This suggests that an ADL-based taxonomy can help guide the development of control subroutines tailored to those specific requirements and that the composition of such subroutines will be capable of solving a broad variety of tasks.
\begin{figure}
    \begin{center}
        \includegraphics[width=0.51\textwidth]{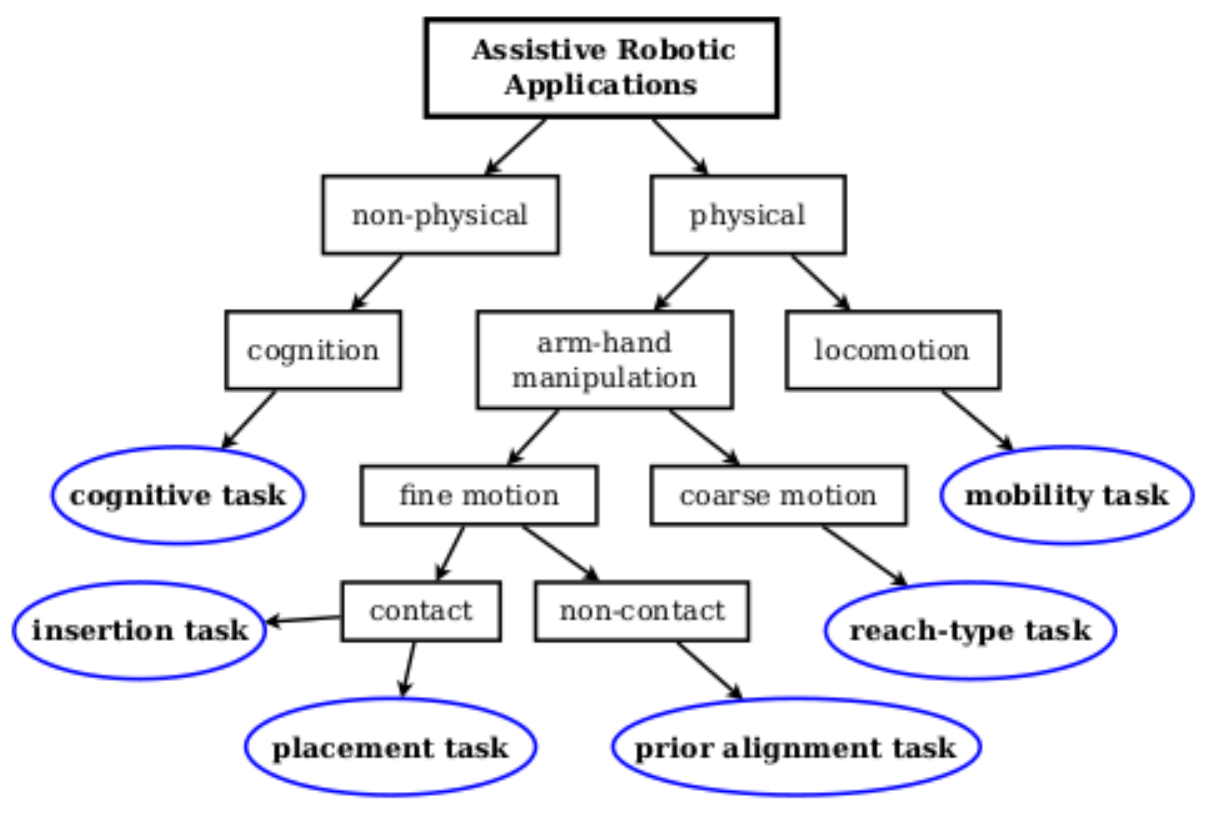} 
        \caption{High-Level Taxonomy of Assistive Robot Tasks and Motions}
        \label{fig:RoboticsTax} 
    \end{center}
\end{figure}

Figure \ref{fig:RoboticsTax} introduces a high-level taxonomy of assistive robotics tasks and arm manipulations. There are three general categories in assistive robotic applications: non-physical cognitive tasks, locomotion-based mobility tasks and arm-hand manipulation tasks. In this work we will focus on analyzing arm and hand manipulations. In applied robotics, the robot gripper is typically used for grasping while the arm is responsible for gross pose alignment and contact point decisions. There is much work to be done before robotic systems will be able to utilize fine dexterous finger manipulation in applied ADLs, with the arm capable of performing both coarse and fine manipulations \cite{Bullock201105Behavior}.
Coarse reaching motions are mostly a 3-DOF translation and requires moderate accuracy. Fine motions can be further subdivided into contact and non-contact depending on the desired outcome. Non-contact 6-DOF fine motions can be used to bring an object into alignment with the target location before putting the object down or inserting it. Although most applied robotics is performed using position-based control, some studies take contact forces into account, either through impedance control or sensing and modeling of the surface for hybrid position-force control \cite{HybridVisionForce}. Surface contact data allows for human-like control strategies to overcome sensing and actuation inaccuracies by utilizing practices such as feeling a table and sliding across it before picking up a small object.

\section{ADL Evaluation from Lifelogging Data}
        \label{sec:lifelogging}
Lifelogging data is a valuable source of quantitative ADL and human motion information. Lifelogging involves long-term recording of all activities performed by an individual throughout the course of the day, usually through a video camera, and occasionally using other types of sensors \cite{GurrinCathalJohoHideoHopfgartnerFrankZhouLitingAlbatal201623Research}. While lifelogging research has been published for over two decades \cite{Mann1997}, hardware and method innovation has made the field grow greatly within the past few years \cite{Bolanos201725Overview}. Small, wearable cameras, such as the Microsoft Lifecam \cite{Wilson2016}, with a longer recording duration has made it more practical compared to the analog video cameras and recorders used in initial research. 
New methods for recognizing objects and actions has driven Computer Vision (CV) research interests to explore lifelogging data, which has been found to be a source of more realistic “in-the-wild” data than typical CV benchmarks \cite{Pirsiavash201214Views, Fathi201122Activities}.

In this work we studied over 30 lifelogging datasets, most of which targeted the performance of a particular algorithm (e.g. video object recognition in home environments) and therefore did not encompass the full day. These datasets typically did not have a statistically sound sampling over all objects and tasks in order to meet our analysis inclusion criteria for this work. We found that long term recordings of several days or more were done at 1-2 frames per minute (fpm), making these useful to analyze gross ADL task frequency and duration, but not suitable for studying detailed timing of individual arm and hand motions. An additional downfall of these datasets is that they fail to capture daily tasks which are repeated with high frequency but are performed quickly, such as opening doors or turning on lights. Another category of datasets had regular video rate recordings of specific tasks, at 30 frames per second (fps), making the detailed timings of individual arm and hand motions possible. We were able to choose three sources of data for analysis: two from long duration recordings in order to extract ADL task frequency and duration \cite{GurrinCathalJohoHideoHopfgartnerFrankZhouLitingAlbatal201623Research,Tapia200446Sensors}, and 
one from short-term recordings of individual tasks \cite{Fathi201221Gaze}.

\begin{figure}
    \begin{center}
        \includegraphics[width=0.5\textwidth]{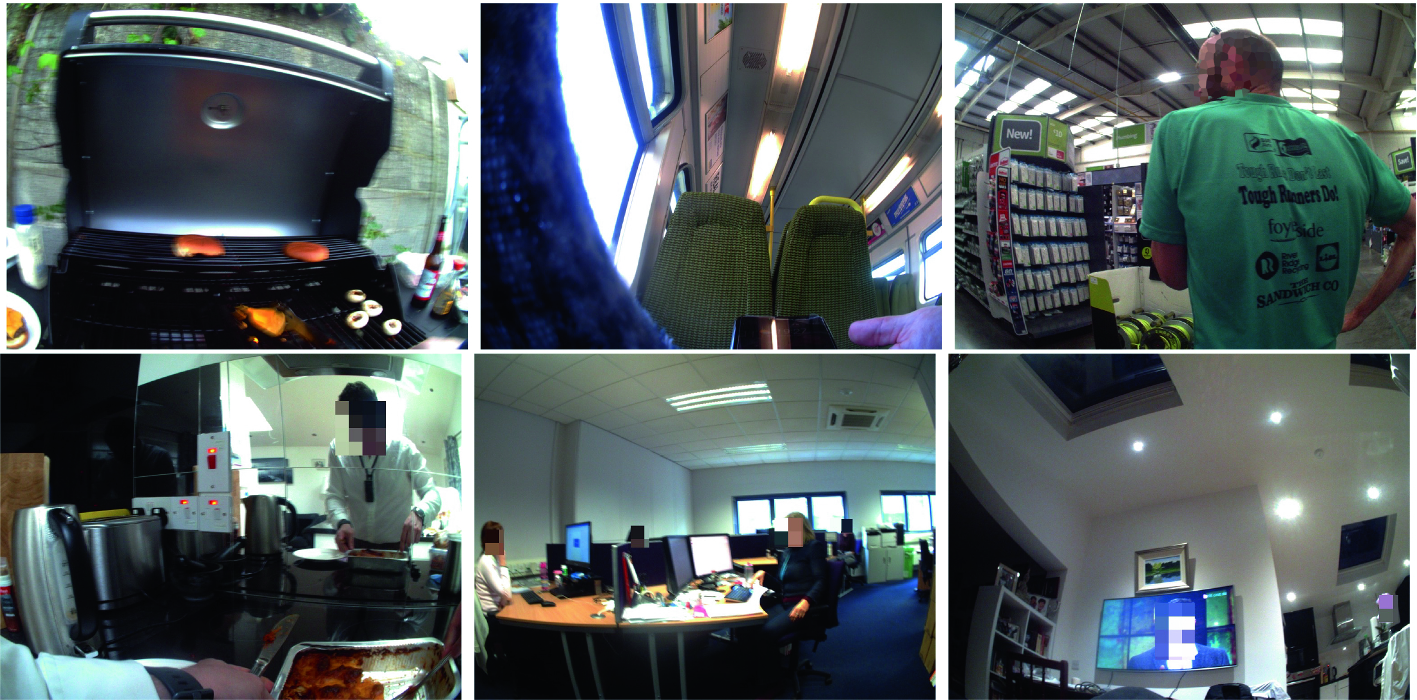} 
        \caption{In the NCTIR Lifelog Dataset \cite{Gurrin201724Task} 3 people wore lifelogging cameras for a total of 79 days, collectively. These provide images of the individuals arms and egocentric environment at a rate of 2 fpm. Due to the low frame rate, fine arm and hand motions are not available, but actions are instead inferred from context using visual concepts automatically computed from the images.}
        \label{fig:NCTIRimages} 
    \end{center}
\end{figure}

\subsection{ADL Task Timing Analysis}
\begin{figure}
    \begin{center}
        \includegraphics[width=0.5\textwidth]{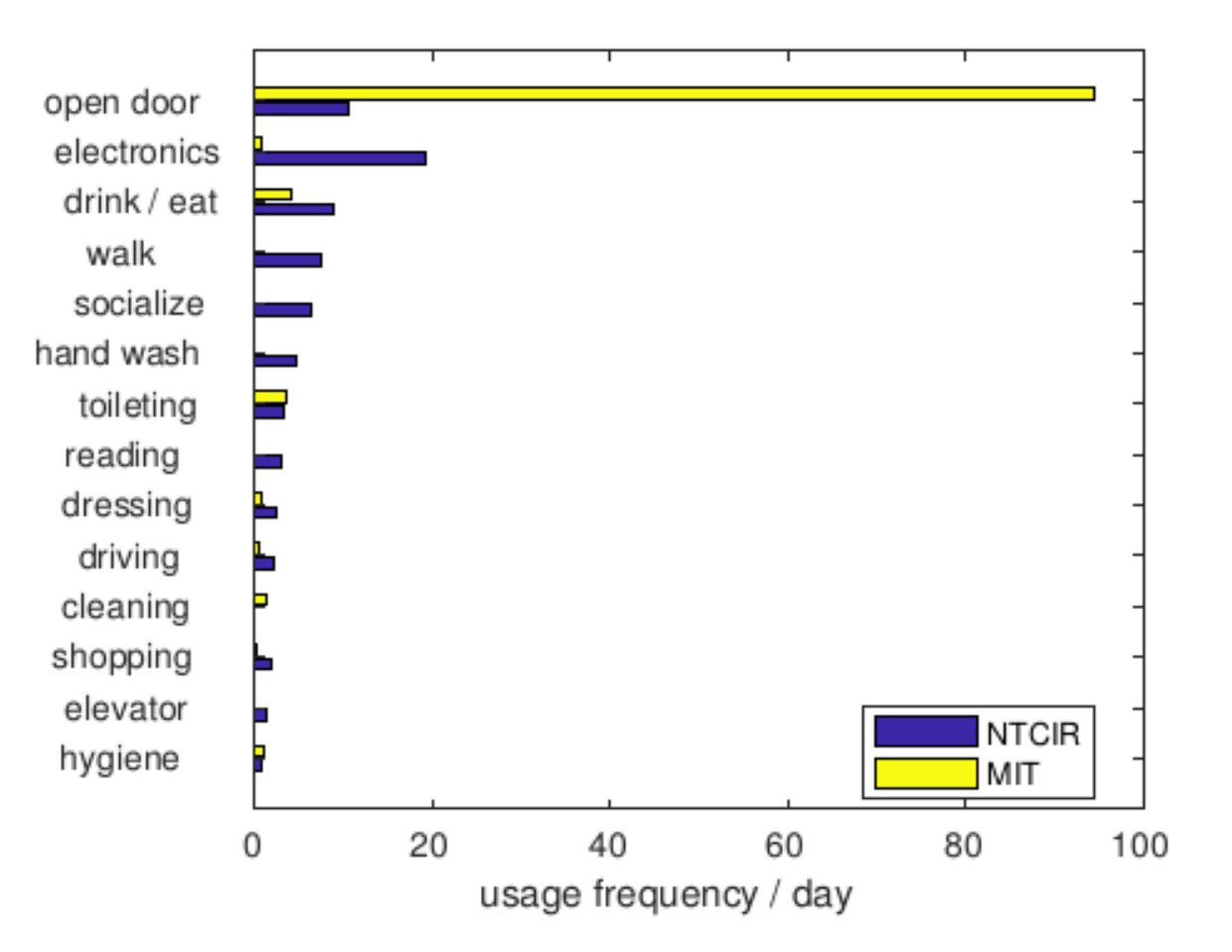} 
        \caption{Human ADL task frequencies from MIT IoT sensors (yellow bars), and NTCIR lifelogging video (blue bars).}
        \label{fig:ADLFreq} 
    \end{center}
\end{figure}

To compute quantitative data on ADL task frequency and duration we analyzed both egocentric lifelogging videos (referred to as ‘NTCIR’ \cite{GurrinCathalJohoHideoHopfgartnerFrankZhouLitingAlbatal201623Research, Gurrin201724Task}), and exocentric data from Internet-of-Things type sensing built into home objects (referred to as ‘MIT’) \cite{Tapia200446Sensors}. Example lifelogging images from the NTCIR dataset are shown in Fig. \ref{fig:NCTIRimages}. The use of complementary sensing turned out to be important for capturing a broader set of tasks. Similar to other CV research, we were able to infer actions from automatically computed visual concepts \cite{Fathi201221Gaze}. Our supplementary web page contains the visual context to actions inference bindings, so readers can replicate results or add other rules and actions to classify. We hand-labeled a small portion of the data to verify the accuracy of the automatic computations. This enabled us to label in-home data sequences spanning multiple days according to what ADLs were carried out at particular times and compute their statistics. Figure \ref{fig:ADLFreq} illustrates the frequency of the most common ADL tasks found in these datasets.

We have grouped tasks together that correspond with robot skills rather than specific healthcare ADL/ICF codes. Some events are detected more reliably by the embedded sensors used in MIT, others only in the lifelogging videos. For examples sensors detect quick events more reliably that the lifelogging video data misses. In contrast, outdoor activities are only captured in the video data. By combining results from both datasets, we were able to obtain an accurate quantitative measure of task significance.

Opening and closing doors is the most frequent task at 94 times per day; this category includes doors between rooms, cabinet doors and drawers. Our rationale for including cabinet doors and drawers is that the robot would approach each situation in the same fashion as a standard door. We believe the MIT data was more accurate since the ‘door opening’ data was obtained from built in door sensors; the low video frequency (2 fpm) of the NTCIR data presented low accuracy with the automatic visual concepts extraction by missing quick openings, particularly of cabinet doors and drawers to retrieve objects. Following door opening, electronics is the second most frequent task performed during the day; the electronics category refers to the use of electronic handheld devices and was dominated by smart phone use. These devices were mostly not covered by the MIT sensors, but were detected in the NTCIR video data. Drinking and eating were essential tasks in both studies, with a frequency of 8.8/day from NTCIR and 4.4/day from MIT. MIT-data captured hand washing by the number of faucet openings/closing (ie. turning the sink on and off resulted in two tasks), which overestimated hand washing frequency. We removed this outlier and relied on the NTCIR results of 4.7/day.



\subsection{Arm and Hand Motion Analysis.}
\begin{figure}
    \begin{center}
        \includegraphics[width=0.5\textwidth]{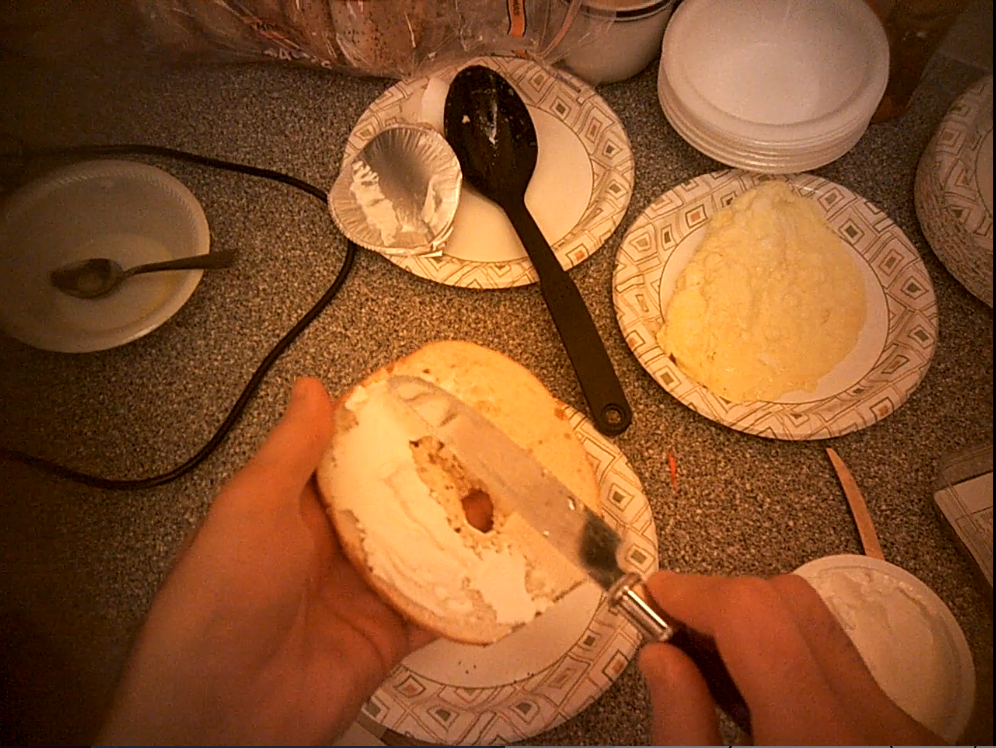} 
        \caption{The GTEA+ dataset contains 7 kitchen activities performed by 6 persons. Image from "American Breakfast"}
        \label{fig:GTEAimage}
    \end{center}
\end{figure}

From other video datasets we were able to extract the number and timings of individual arm and hand motions required to perform a particular ADL and, for a few tasks, similar timings for robot execution. The Georgia Tech Egocentric Activity Datasets (GTEA+) \footnote{http://www.cbi.gatech.edu/fpv/} contain full frame rate (30 fps) video recordings of humans performing domestic tasks \cite{Fathi201221Gaze}. We analyzed the annotated GTEA+ dataset, which contained approximately 25GB of annotated kitchen activity videos to extract individual human motion timings performed during these tasks (Fig. \ref{fig:GTEAimage}).

\begin{figure}
    \begin{center}
        \includegraphics[width=0.5\textwidth]{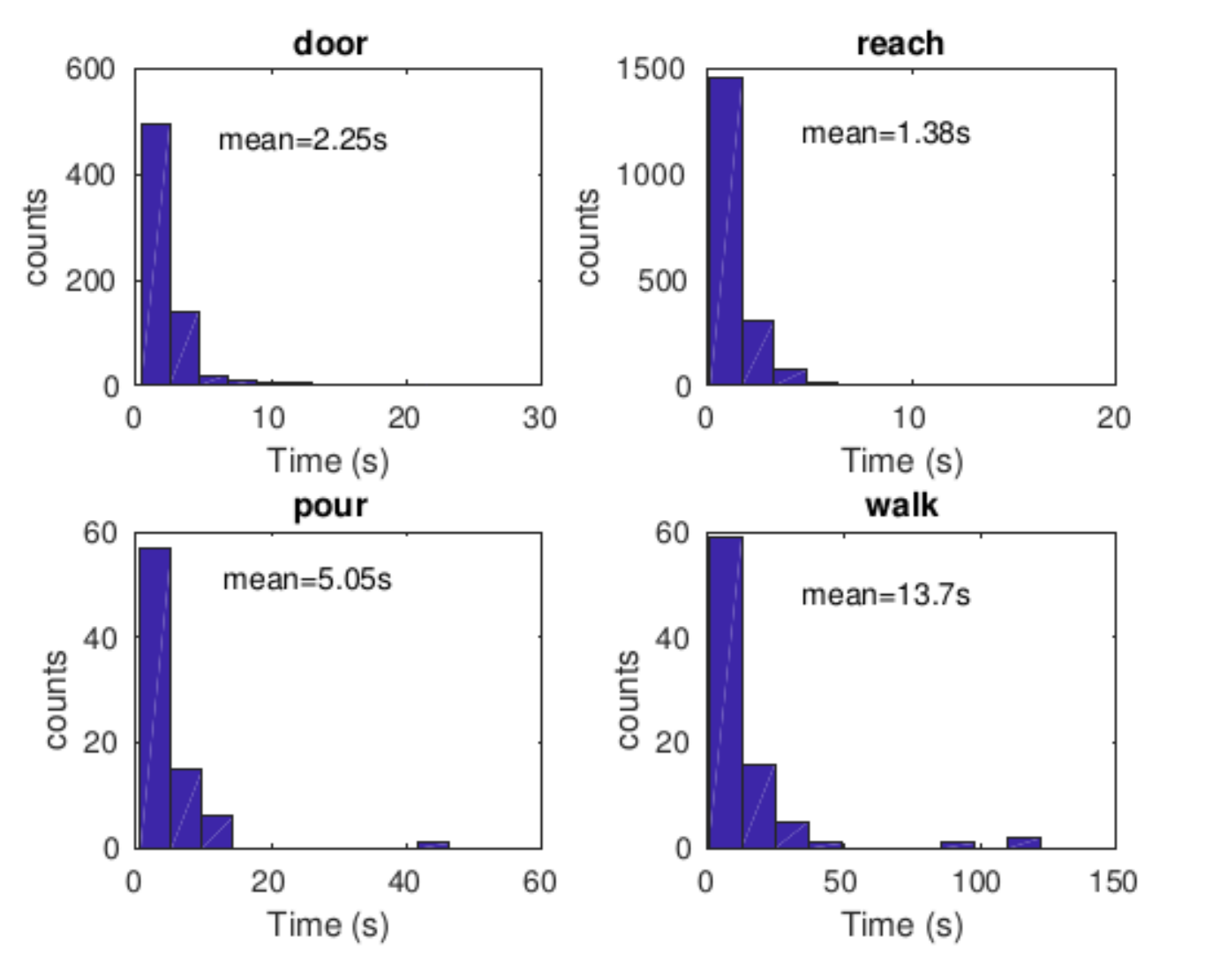} 
        \caption{Timing histograms for four common human motions}
        \label{fig:GTEAmotions} 
    \end{center}
\end{figure}

Figure \ref{fig:GTEAmotions} illustrates four common motions out of the 33 in GTEA+. Notably, human motions were far faster than typical assistive robot motions. For example, as seen in the histogram, reach motions that take just a second for a human, can take anywhere from ten seconds to several minutes in published HRI solutions  \cite{Muelling201547Manipulation}. This has implications for how many human motions and ADLs a robot system can practically substitute in a day without taking up an excessive amount of time. In other motions, such as pouring liquids, the task itself constrains the human to proceed rather slowly. The door task covers both lightweight cabinet doors and drawers, along with heavier doors (e.g. refrigerator); with lighter doors, the human times approached that of an unconstrained reach, despite the more challenging physical constraint of hinged or sliding motion, while heavier doors represent the long tail of the time distribution. Unlike NTCIR, GTEA+ is not a representative sampling of all human activities. It is still notable that the number of reaches is three times the number of door openings (1500 reaches versus 500 door and drawer openings over 11 hours of video). 

We have presented only a selection of human motions and activities here to support our discussion and conclusions. Statistics on the full set of human Activities of Daily Living and human motion data are available on our companion web site: \url{http://webdocs.cs.ualberta.ca/~vis/ADL/}.

\section{Discussion}
\label{sec:discussion}
Door opening/closing, drinking/eating, hand washing and toileting would arguably be the most essential to support for assistive robot arm and hand systems, out of all the ADL tasks analyzed in this work. The first three are relatively feasible to accomplish given the payload capacity of current robotic arms. 

Activities such as using electronics (primarily smartphones), socializing, and reading could be physically aided by robot arms, but since these activities are not inherently physical, alternative solutions are possible and can be a simpler and more reliable solution (e.g. hands-free phone use and other computational automation).

Toileting is a high priority task that involves transferring from a wheelchair to the toilet. Assistive arms do not support this, but there are specialized transfer devices - also useful for transfer from beds - that are generally used in health care, and be employed in peoples homes.

Overall, there is great potential for supporting ADLs for those living with disabilities as well as the elderly. Over the past few decades there has been an increasing demand for health care services due to the rising elderly and disability populations \cite{DepartmentofEconomicandSocialAffairs:PopulationDivision2017WorldHighlights}. Assistive robots can help bridge this gap by alleviating the labour burden for health care specialists and caregivers. Furthermore, an assistive robot could help one perform ADL they are otherwise incapable of managing on their own, increasing functional independence. 

However, challenges remain before these robots will reach mainstream adoption, including but not limited to: system costs, task completion times and ease of use via user interfaces. Currently costing around USD 30,000, an arm is a significant expense for an individual, who may already have a limited income. While western health insurance often covers expensive prostheses for amputees, only in the Netherlands does insurance cover a wheelchair mounted arm. The hardware of an arm consists of motors, gears, links and computing, all of which in principle can be bought and integrated inexpensively in volume production. However, to reach a high enough volume, health providers in major countries need to come on board. These are in turn are waiting for lower costs, and more in-the-field proof of benefits. Fitting the robot and often a custom user interface for the particular individual is another varying additional cost. For a user who has reduced upper arm movements, but maintains precise control, a standard joystick can be used, and a standard robot-to-wheelchair fitting kit costs approximately 1000 USD. In more complicated cases, fitting costs can be substantially more.

Speed of robot motion, which affects task completion time, is another challenge. While a human reach takes on average 1.4s (Fig.~\ref{fig:GTEAmotions}), published assistive robots take 40-90s, resulting in robot solutions that are nearly 10 times slower \cite{Kim201252Robot,QuinteroRJ15,Muelling201547Manipulation}. In the GTEA kitchen tasks, users performed 150 reaches per hour. Substituting robot reaches for humans would turn a moderate 30 minute meal preparation and eating time into a 5 hour ordeal. Anecdotal comments from users of assistive robot arms are that everyday morning kitchen and bathroom activities, which an able person easily performs in less than an hour, takes them several hours.

Robots may solve tasks differently than humans as robots are often limited to grasping one item at a time, while humans can handle many. When setting a table we will for instance pick several utensils at a time from a drawer. In restaurants, waiters can clear a table for four, and handle all the plates, utensils, glasses etc in their hands and arms. Analysing the publicly available TUM Kitchen Data Set of activity sequences recorded in a kitchen environment \cite{Tenorth2009}, we found that the robot strategy on average required 1.6 times more movements than a human. Users of assistive robots adopt compromises to deal with the speed and accuracy of robots. For example, foods and drinks that can be consumed while held statically in front of the humans face by the robot, e.g. eating a snack bar, or drinking with a straw, are far quicker to consume than those requiring numerous robot reach motions, such as eating a bowl of cereal.
 
User interfaces need improvements. Currently deployed arms are, as mentioned before, joystick operated, while most research is on autonomous movement, e.g. autonomously delivering a piece of food once the system has detected an open mouth \cite{Park2019, Gordon2019}. Sheridan's conventional scale from tele-operation to autonomy \cite{Sheridan2002}, has been redefined by Goodrich to have human-robot collaboration as the goal rather than robot autonomy \cite{Goodrich200706bSurvey}.

We and others have found that users generally prefer to have continuous in-the-loop control\cite{QuinteroRJ15, Kim2012}. Someone may change their mind midway through autonomous food delivery, and may instead open their mouth to say something - only to get their mouth stuffed with food. In very recent work a low dimensional control space is learned from demonstrations. This allows a human user to have direct control over a 6DOF motion using a low DOF HRI, such as a joystick \cite{Jeon2020, Quintero2017}. Getting the balance right between human interaction and semi-autonomous assistive systems will be challenging. Currently, most research is evaluated with a few participants trying it for about an hour each in a research lab setting. We expect that new HRI solutions will need to be deployed longer term in real users homes in order to properly evaluate usability. 

What is the magnitude of need and potential for robotic assistance in the word? Many countries in the west and Asia have aging populations and disabilities also affect younger populations, e.g. from car accidents, disease or inheritance. Definitions and quality of statistics on disability differs across nations, and are difficult to integrate globally. 
Canada has a multi-ethnic population and characteristics similar to many other industrialized nations. The proportion of seniors (age 65+) in the population is steadily increasing, with seniors comprising a projected 23.1\% of the population by 2031 \cite{StatCan2}. In 2014, seniors constituted only 14\% of the population, but consumed 46\% of provincial public health care dollars \cite{StatCan3}. A growing number of elderly and disabled, supported by a dwindling young population is putting pressure both on government budgets and available health care personnel. Today, individuals with lower body impairments and the elderly are able to independently move around using power wheelchairs. In the near future wheelchair mounted robot arms could help increase independence and reduce care need for those living with reduced upper limb function.

Statistics Canada found that from 2001 - 2006 there was a 20.5\% increase in those identifying as having a disability, corresponding to over 2.4 million people in Canada \cite{StatCan}.  
One in twenty Canadians living with disabilities regularly receive assistance with at least one ADL on a daily basis, although not all of which will require the use of wheelchair mounted arms. This suggests that there is a significant need and potential market for robotic solutions in Canada and similar countries across the world. Some may prefer automation integration with their smart homes, and some may require both cognitive and physical assistance. While artificial intelligence might provide some basic cognitive support, planning of the days tasks and reminders, it cannot eliminate the need for human contact and support. However, robotic assistance can free up humans from more mundane chores, allowing more time for caregivers to focus on high quality help and personal interaction. 

An advantage of wheelchair mounted arms is that they are with the person at all times, both in home or when out and about, while nursing care is typically only for the morning and evening routines for those who live independently. Image dropping something important and having to wait all day before someone is able to help retrieve it.


\section{Conclusions}
In this paper we presented assistive robotics for Activities of Daily Living both from a health care perspective and robotics perspective. We analyzed human ADL task frequency from public life-logging datasets and computed motion timings from public Computer Vision data. Overall, reach motions (to grasp objects) and door openings (including cabinets and drawers) were the most frequent motions. Drinking, eating and hand washing are other high priority tasks that can be addressed by current assistive robot arms. Toileting and dressing, while ranking just below, are generally thought to be more challenging for robotics, since they require the transfer of body weight. Detailed data on frequency and duration information for all analyzed tasks and motions, as well as the analysis methods are available on the companion website \url{http://webdocs.cs.ualberta.ca/~vis/ADL}
\label{sec:conclusions}


\newpage
\bibliographystyle{IEEEtran}
\bibliography{References}
\addtolength{\textheight}{-12cm}
\end{document}